\documentclass[11pt,a4paper]{article}
\usepackage{lscape}
\usepackage{amsfonts, amsmath,amssymb}
\usepackage{graphicx}
\usepackage{graphics}
\graphicspath{ {./Bilder/} }
\usepackage{caption}
\usepackage[font=scriptsize]{caption}
\usepackage{epstopdf}
\usepackage{amsthm}
\usepackage{url}
\usepackage{ucs}
\usepackage[T1]{fontenc}\usepackage[utf8]{inputenc}
\usepackage{tabularx}
\usepackage{longtable}
\usepackage{ltablex}
\usepackage{hyperref}
\usepackage{color}
\usepackage{rotating}
\usepackage{lscape}
\usepackage{supertabular} 
\usepackage{natbib}
\begin{document}

\baselineskip=1.3\normalbaselineskip

\title{BERT based freedom to operate patent analysis}

\author {Michael Freunek \thanks{University of Bern, Institute of Mathematics, Sidlerstrasse 5, 3012 Bern, Switzerland, Email: \texttt{michael.freunek@math.unibe.ch, m.freunek@gmx.de}} \and Andr\'e Bodmer \thanks{University of Berne, Institute of Economics, Schanzeneckstrasse 1, 3001 Bern, Switzerland, Email: \texttt{bodmerandre.ab@gmail.com, andre.bodmer@vwi.unibe.ch	}}}

\date{4. May 2021}
\maketitle

\begin{abstract}
\noindent
In this paper we present a method to apply BERT to freedom to operate patent analysis and patent searches. According to the method, BERT is fine-tuned by training patent descriptions to the independent claims. Each description represents an invention which is protected by the corresponding claims. Such a trained BERT could be able to identify or order freedom to operate relevant patents based on a short description of an invention or product.
We tested the method by training BERT on the patent class G06T1/00 and applied the trained BERT on five inventions classified in G06T1/60,  described via DOCDB abstracts. The DOCDB abstract are available on ESPACENET of the European Patent Office.

\bigskip


\end{abstract}

\maketitle

\clearpage

\section{Introduction}

\noindent
With new inventions and the launch of new products, an inventor or entrepreneur is often faced with the question of possible patent infringement. There are risks of high procedural costs, expensive penalties and compensation as well as a ban on the sale of the product, despite high development expenditures. Therefore, inventors try to assess the risks of patent infringement before commencing commercial activities, which can possibly be associated with an extensive freedom to operate (FTO) analysis.

FTO patent analysis are generally very time-consuming. Patent claims must be identified, which could cover the invention or product. It should be taken into account, that the claims of pending patent applications can still change during the examination process. Often the claims become modified by adding further technical features. 

Currently, the vast majority of patent searches are performed based on Boolean methods. Patents are preselected on the basis of keywords, usually in combination with suitable IPC (International Patent Classification\footnote{\url{https://www.wipo.int/classifications/ipc/en}.}) and CPC (Cooperative Patent Classification\footnote{\url{https://worldwide.espacenet.com/classification?locale=en_EP}.}) patent classes. Nevertheless, usually hundreds to more than a thousand patent documents remain that have to be viewed and assessed by a patent expert. Even with only a small amount of time spent per patent, the total effort is enormous and it is not uncommon for relevant patents to be overlooked. With a Boolean method, all synonyms must also be considered in the keyword concepts, which can be very difficult, especially in patent language. Also the relationship of the keywords to one another, which have to be modeled in a complex manner with distance operators, can lead to lose interesting patents. Heteronyms can lead to a large number of irrelevant documents, which makes evaluation by the patent expert more extensive and error-prone. The work is also becoming more and more complex due to the rapidly growing number of new patent applications and increasing number of technological interdisciplinary inventions.

In this paper we try out a method to support FTO patent searches and analysis with the help of machine learning and to improve the efficiency of the search and patent evaluation. 

This work essentially reverses the procedure described in the publication \cite{freunek2021bert}, where BERT is applied to patent novelty searches. While a relevant description is searched for a claim as input in \cite{freunek2021bert}, here, BERT is fine-tuned for the task to identify FTO relevant independent claims for a description of an invention or a product.

For this reason, we concatenate descriptions to their independent claims. Each description represents an invention, which is protected by the corresponding claims. The idea is, by fine-tuning BERT on that task, BERT learns to identify FTO relevant claims based on a description of an invention or a product as input.

In this paper we describe this method in detail and as a test, apply BERT to five DOCDC abstracts of inventions. To our knowledge, the method described in this work has not been published yet. The method presented here offers the following advantages (Freunek, Bodmer):

\begin{itemize}
\item Data availability and data volume: Basically every patent  (in an appropriate language) is suitable for training. Almost any amount of data is available for any technology. This means that BERT can be trained very precisely for any searched description.
\item Independence from published data sets. 
\item The handling of the described procedure is also quite simple and straightforward. 
\end{itemize}

\noindent
The paper is structured as follows: First, we identify related work in Chapter \ref{related work}. Then we show our approach to train BERT on FTO analysis in Chapter \ref{approach}: How we generate the data and which BERT model we use, how we practically apply the trained BERT for FTO patent searches or analysis. In Chapter \ref{experiment}, we will apply the described method and train BERT on real patent data and finally test the trained BERT on five inventions to identify FTO relevant patents. In Chapter \ref{conclusion} we give some conclusions and possibilities for future research tasks.\\

\section{Related Work}
\label{related work}

\noindent
This paper mainly refers to the work in \cite{freunek2021bert}. There, the idea is to concatenate the patent claims to their own description to fine-tune BERT for patent novelty searches. Such a trained BERT should be able to identify novelty destroying descriptions for claims. This work is very closely related to this idea, hence this paper shows many parallels to the cited work.

BERT is a pre-trained model for tasks in natural language processing and is based on a transformer architecture. The self-attention mechanism relates different positions in a sentence or input sequence to each other, as described in \cite{Vaswani2017} and \cite{Devlin2018}. The pre-trained BERT is fine-tuned to the task of interest.

Some papers describe the application of BERT in the field of patent classification like in \cite{Sun2019} and \cite{Lee2020}. In addition to the work in \cite{freunek2021bert}, the paper \cite{Risch2020} also describes the use of BERT for novelty searches. A detailed study on patent novelty detection can be found in \cite{Chikkamath2020AnES}. The authors achieved their best results NBSVM (Naive Bayes Support Vector Machine). A further study about prior art searches in combination with GPT2 can be found in \cite{Lee2020a}.

A survey about deep learning methods applied to patent analysis is given in \cite{KRESTEL2021102035}. Among other things, the authors identified the trend towards more complex models such as BERT. In the work of \cite{Setchi2021}, a platform for comparison of state of the art AI techniques in patent searches are demonstrated. A further literature overview of applying AI to intellectual property can be found in \cite{Aristodemou2018} or the application to patent landscaping has been reported in \cite{Choi2019}.

The authors of this paper are not aware of any publications focused on FTO patent searches based on machine learning methods. Possibly, methods that deal with the search of prior art could also be applied to the subject of FTO analysis. 
\\

\section{Approach}
\label{approach}

The work is based on the following idea: With the development of an invention or a product, the question of its commercial usability arises, whether the invention or the product is already protected by a patent. We will describe how to train BERT on this question. The invention or the product can be illustrated by a brief description. Based on this brief description, the relevant patents are to be identified with the help of BERT.
In each patent, there is a description for an invention, which is covered by the corresponding claims. Therefore, in this work we will link the descriptions in the patents with their own independent claims. BERT should thus learn relevant claims to descriptions and finally be able to identify claims or patents FTO relevant to a description as input. \\

\subsection{Data Generation}
\label{DG}

To generate the training data for BERT one has to be aware of two things: 1: BERT can process input lengths up to 512 tokens. 2: We suppose that a suitable description of an invention or product is of size of a typical patent abstract, which has a length of roughly 200 words. Since a patent description can amount to several thousand words, we have to slice the description into pieces, which should have the typical length of a patent abstract and, concatenated to an independent claim, fits the input size of BERT with respect to special tokens, like the CLS (for classification) and the SEP (for separation) token. The procedure of slicing patent descriptions is explained in some detail in \cite{freunek2021bert}. We do not select specific text passages from the description, but rather link the entire description to the independent claims. Of course, the concatenated claim will not always match the item of description piece. For example, when the description cites the related prior art and its disadvantages. However, the procedure described should be a good approximation from a statistical point of view, since the majority of the description in a patent is directly dedicated to the actual invention and its embodiments. 

In more detail, this works as follows: Each description-piece $k$ of patent $i$ is combined with the independent claim(s) of the same patent $i$ yielding the first label data: 

\begin{equation*}
\begin{split}
\textrm{Input}_{i,k} = \textrm{description-piece}({\textrm{patent}_i})_k \textrm{ <> } \textrm{SEP} \textrm{ <> }  \textrm{claim}({\textrm{patent}_i})
\end{split}
\end{equation*}
\\
Now, having the data for the first label, we need to complete the training data set by generating data with the second label, where is no matching between description piece and independent claim, or in other words, where the claim does not cover the description and thus is of no FTO relevance. We make use of the observation , that most independent claims of patent $i$ are of no FTO relevance of the description of patent $j$, with $i \ne j$. Applying this observation, we can generate the second label data according to:

\begin{equation*}
\begin{split}
\textrm{Input}_{i,j, k} =  \textrm{description-piece}({\textrm{patent}_j})_k \textrm{ <> } \textrm{SEP} \textrm{ <> } \textrm{claim}({\textrm{patent}_i}),
\end{split}
\end{equation*}
\\
where $i \ne j$. Description pieces and the independent claims are combined in a random manner. The only limitation is that claim and description piece belong to different patents.

By taking the same claims and descriptions pieces for the second label as for the first label, one can make sure, that both labels have the same frequency and BERT learns to distinguish between relevant and non-relevant claims for patent descriptions. Of course, there is a probability of concatenating a description piece randomly to a FTO relevant claim of another patent, but at the end the effect a "confusing" during training should be very low. Finally, each input has the structure:

\begin{equation}
\textrm{CLS} \textrm{ <> } \textrm{description} \textrm{ <> } \textrm{SEP} \textrm{ <> } \textrm{claim} \textrm{ <> } \textrm{PAD}\textrm{ <> }\textrm{SEP} \textrm{ <> }\textrm{ <> } \textrm{label},
\label{structure1}
\end{equation}
\\
where the length of the input is padded with PAD tokens during tokenization to the defined maximum input length, if the length of the concatenated description piece and claim is smaller than the maximum input length. Otherwise we get:

\begin{equation}
\textrm{CLS} \textrm{ <> } \textrm{description} \textrm{ <> } \textrm{SEP} \textrm{ <> } \textrm{claim} \textrm{ <> } \textrm{SEP} \textrm{ <> }\textrm{ <> } \textrm{label}.
\label{structure2}
\end{equation}
\\
The maximum input length can be set to lower values than 512 tokens, but it is expected to get the best results, when the description pieces and claim concatenations fit the maximum input size of BERT.
\\

\subsection{BERT Model and Training}
\label{bert_model_training}

For training BERT, we applied the tokenizer {\bf bert-base-uncased}. An important parameter here is the maximum input sequence length. BERT allows a maximum input length of 512 tokens. It is intuitively obvious to go close to this limit with this model, provided that the computer capacity allows it. In our experiment in chapter \ref{experiment} we set the input length to 500 tokens.

Hyperparameters like learning rate, batch size, number of epochs have to be adjusted and a BERT model has to be selected. The probably model of choice is the {\bf Bert For Next Sentence Prediction}. As in \cite{freunek2021bert}, we will chose the model {\bf Bert For Sequence Classification}.

\subsection{Applying Trained BERT to Freedom to Operate Patent Analysis}
\label{applying_trained_bert}

Once BERT is trained (fine-tuned), BERT can be applied to FTO patent analysis or patent searches with a short description of an invention or a product as input. In our studies we focused training BERT on selected technology fields as in \cite{freunek2021bert}. The required computer capacity and the volume of data is therefore significantly lower. Also, it cannot be ruled out that a technology-specialized BERT is superior to a general trained BERT (trained to all or at least to several technology fields). The disadvantage, however, is that BERT must be trained specifically for a task, provided that the technology has not been trained before.

To prepare the input, we prepare the patents similar to the data generation procedure described in chapter \ref{DG}: The independent claims, which we want to analyze according to FTO relevance for the description of interest, are concatenated according to the structure ("equation") \ref{structure1} or \ref{structure2}. The concatenations should fit the input size of BERT without slicing, since descriptions of inventions or products have typically the size of an abstract and typical abstract and claim concatenations have lengths below 512 words.

To be clear: We train BERT on the complete description of the patents. But the FTO analysis is carried out for a short description of an invention or product, which has the size of an abstract.

Then, the description of the invention or product - $\textrm{description}_\textrm{ip}$ - is concatenated to the independent claims, we get the following structure (by neglecting the CLS and PAD token):
\begin{align*}
...\\
\textrm{Input}_{m}   &= \textrm{description}_\textrm{ip}\textrm{ <> } \textrm{SEP} \textrm{ <> } \textrm{claim}({\textrm{patent}_{m}}) \\
\textrm{Input}_{m+1} &= \textrm{description}_\textrm{ip}\textrm{ <> } \textrm{SEP} \textrm{ <> } \textrm{claim}({\textrm{patent}_{m+1}}) \\
\textrm{Input}_{m+2} &= \textrm{description}_\textrm{ip}\textrm{ <> } \textrm{SEP} \textrm{ <> } \textrm{claim}({\textrm{patent}_{m+2}}) \\
...
\end{align*}

\subsection{Evaluation of the BERT Results}

According to chapter \ref{applying_trained_bert}, BERT assigns to the analyzed claims logits numbers for both label. This logits represent the assessment of the claims of being FTO relevant (first label) or being non-relevant (second label). If desired (but not necessary), the logits can be further calculated with the Softmax function. The claims should be arranged according to the values of the logits or the Softmax function according to the first label. This results in a sequence according to the relevance of the claims according to BERT´s calculation.

\section{Experiment}
\label{experiment}

\noindent
Basically we are faced with the problem, that there is no reference data set and no strict evaluation scheme to assess the performance of BERT or another machine learning model on a FTO task. For a classification task, the precision, the recall and the F1 score could be calculated. This is hardly possible for a FTO analysis. This experiment is therefore intended to give an initial indication of whether the method described above can be used reasonably.

We applied the method to the DOCDB abstracts of five patents (reference patents). The abstracts should represent a short description of the corresponding invention or product, for which we want to make the FTO patent search or patent analysis. BERT should identify the most FTO relevant claims for the five reference patents.

To this end, we performed the following procedure:

\begin{itemize}
\item[1:] Generation of 1059 training patents of label 0 and label 1 according to chapter \ref{DG} by a random selection of patents, which refer to the same patent class (here on IPC level G06T1/00).
\item[2:] Training BERT with the training patents on the first claims.
\item[3:] Random selection of a second group of 2577 patents (to-be-searched group), chosen from the classification G06T1/60. The five reference patents are included in this to-be-searched group. There is no overlap between the training and the to-be-searched patents. BERT analyzes this to-be-searched group for an FTO analysis based on the DOCDB abstracts of the five reference patents.
\item[4:] Identifying the most relevant claims /patents according to the logits or Softmax values.
\end{itemize}

\subsection{Data}
\label{the_data}

The FTO analysis of the first claims is performed for the following patent applications (reference patents): AU~2012315252~B2, US~9,659,410~B2, \\US~9,911,174~B2, US~10,375379~B2 and US~2015/0302541~A1.

All 2577 to-be-searched patents, including the reference patents, are assigned to the IPC or CPC patent class G06T1/60.
The 1059 training patents are randomly selected from G06T1/00 and below. All patents are US, GB, AU, CA or IN patents in English language. It was checked, that BERT was not trained on the reference and the other to-be-searched patents, although this is not a condition or limitation of the method presented. Between the randomly chosen training patents and the randomly chosen to-be-searched patents is no overlap. The descriptions of the patents were sliced into pieces of random size in the range of 100 to 200 words. Future research could investigate a dynamic slicing to fit exactly the size of the concatenated claim and description piece to the maximum sequence length, as described in chapter \ref{DG}. The maximum sequence length for BERT has been chosen to 500 tokens. Slicing the descriptions of the training patents and the generation of label 0 and label 1 description pieces and claims according to chapter \ref{DG}, yields 74'498 training input sequences. We split randomly 9'250 concatenated claim and description pieces for validation.
\\

\subsection{Results: BERT's Freedom to Operate Patent Analysis}
\label{test_bert_novelty_search}

We trained BERT for 2 epochs. The FTO patent search or patent analysis in the to-be-searched group with 2577 patents was performed only on the first claim. The trained BERT yields the results shown in the tables \ref{reference_patent1} - \ref{reference_patent5}. The abstracts and claims of the patents are listed in detail in the Appendix. The results are discussed in the following section.\\

\begin{table}[ht]
\centering
\begin{tabular}[t]{|l|l|}
\hline
\bf{Reference patent 1} & \bf{Pos. / FTO-patent} \\
\hline
AU2012315252B2 & 1.~~AU2012315252B2  \\
& 2.~~US10186068B2   \\ 
& 3.~~US20200058273A1  \\
& 4.~~US10089711B2 \\
& 5.~~US9501415B1 \\
& 6.~~US10643746B2  \\
& 7.~~US20170243322A1  \\
& 8.~~US10452223B2 \\
& 9.~~US8965933B2  \\
& 10.~US9792881B2 \\
\hline
\end{tabular}
\caption{Reference patent 1 is given in the first column, the FTO relevant patents evaluated by BERT are given in column 2.}
\label{reference_patent1}
\end{table}%

\begin{table}[ht]
\centering
\begin{tabular}[t]{|l|l|}
\hline
\bf{Reference patent 2} & \bf{Pos. / FTO-patent} \\
\hline
US9659410B2 & 1.~~US9659410B2 \\
& 2.~~US9666108B2  \\
& 3.~~US10762713B2  \\
& 4.~~US10803826B2  \\
& 5.~~US9978114B2  \\
& 6.~~IN271214B  \\
& 7.~~US10176739B2  \\
& 8.~~US10930084B2 \\
& 9.~~US9626799B2  \\
& 10.~IN201947028682A \\
\hline
\end{tabular}
\caption{Reference patent 2 is given in the first column, the FTO relevant patents evaluated by BERT are given in column 2.}
\label{reference_patent2}
\end{table}%

\begin{table}[ht]
\centering
\begin{tabular}[t]{|l|l|}
\hline
\bf{Reference patent 3} & \bf{Pos. / FTO-patent} \\
\hline
US9911174B2 & 1.~~US9911174B2 \\
& 2.~~US9412147B2  \\
& 3.~~US9898799B2  \\
& 4.~~US10013744B2  \\
& 5.~~IN201934037994A \\
& 6.~~US20200098079A1  \\
& 7.~~US10255655B1   \\
& 8.~~US9996894B2 \\
& 9.~~US20180365796A1  \\
& 10.~GB2469526B \\
\hline
\end{tabular}
\caption{Reference patent 3 is given in the first column, the FTO relevant patents evaluated by BERT are given in column 2.}
\label{reference_patent3}
\end{table}%

\begin{table}[ht]
\centering
\begin{tabular}[t]{|l|l|}
\hline
\bf{Reference patent 4} & \bf{Pos. / FTO-patent} \\
\hline
US10375379B2 & 1.~~US10375379B2 \\
& 2.~~US20150168613A1  \\
& 3.~~US9047839B2  \\
& 4.~~US10204393B2  \\
& 5.~~US9208714B2 \\
& 6.~~US20150287393A1  \\
& 7.~~IN201647038229A   \\
& 8.~~US10522113B2 \\
& 9.~~IN201403134I4  \\
& 10.~US20170076417A1 \\
\hline
\end{tabular}
\caption{Reference patent 4 is given in the first column, the FTO relevant patents evaluated by BERT are given in column 2.}
\label{reference_patent4}
\end{table}%

\begin{table}[ht]
\centering
\begin{tabular}[t]{|l|l|}
\hline
\bf{Reference patent 5} & \bf{Pos. / FTO-patent} \\
\hline
US20150302541A1 & 1.~~US20150302541A1 \\
& 2.~~US9734549B2  \\
& 3.~~US9530177B2 \\
& 4.~~US9451041B2 \\
& 5.~~US20170083450A1 \\
& 6.~~US10719433B2  \\
& 7.~~US9667531B2   \\
& 8.~~CA2813877C \\
& 9.~~US20170140497A1  \\
& 10.~US9153211B1 \\
\hline
\end{tabular}
\caption{Reference patent 5 is given in the first column, the FTO relevant patents evaluated by BERT are given in column 2.}
\label{reference_patent5}
\end{table}%

\subsection{Discussion}
\label{discussion}

The result is difficult to assess. There is neither a standardized data set nor can one clearly assign relevance or non-relevance to each patent. In addition, in the selected patent class G06T1/60, the claims are very long, i.e. have many features. In practice, each of these features can determine whether the patent is FTO relevant or not. An assessment would be much easier if one were to test patents of a patent class which on average require fewer features in the independent claims. Ultimately, the method presented here will have to be tested in practice to assess its efficiency. Apparently the result looks promising.

The first position in the results of the selected FTO patents is noticeable: this is the patent itself, the abstract of which was used as input to BERT. BERT has not previously learned a direct relationship between this abstract and its first claim. The result shows that, among all claims from the 2577 to-be-searched patents, BERT rated the claim to the patent for which we carried out an FTO analysis as the most relevant, without having "seen" the claim beforehand. This result clearly indicates that BERT is able to identify relevant claims, if existing. 

The following points should be noted here, which distinguish the test presented here from a real FTO, but do not represent any restriction with regard to the practicability of the method described: The documents evaluated by BERT may not be complete as other patent classes were completely disregarded. It was also not taken into account whether the patents are still active and it did not distinguish between A-documents (applications) and B-documents (granted patents). Also, only the first claim was evaluated, but not, if existing, further independent claims. 
\\

\section{Conclusion and Future Research}
\label{conclusion}

\noindent
In this work, we presented a new method to train BERT for a freedom to operate (FTO) patent search or patent analysis, by concatenating patent descriptions to their independent claims. The descriptions of the patents are sliced into description pieces of a certain length, which should be adopted to the length of the trained or searched claim. In our tests, we sliced the description into pieces with a random length in the range of 100 to 200 tokens. We applied the trained BERT to a FTO patent search or patent analysis for five inventions, which are described in five DOCDB abstracts. The result is difficult to evaluate, since there is neither a standardized data set nor can one clearly assign relevance or non-relevance to each patent. The apparent result is promising, but the method needs to be further evaluated in practice. 

The following points could be interesting for future research:

\begin{itemize}
\item[1:] Length of the text pieces: We expect the best result when the description pieces are at least as long as the claim or even longer.
\item[2:] We trained BERT particularly on B patents (granted patents). The independent claims of B patents tend to have more features than independent claims of the corresponding applications (A documents). The question arises whether BERT should not be better trained on A patents or a mixture of A and B patents.
\item[3:] We have taken the approach of training BERT technically very close to the features to be searched. However, it may be sufficient here to train the rough technical environment, which would simplify the selection of training patents.
\item[4:] We trained and applied the method only to the first claims. An adoption to further independent claims would be very interesting.
\item[5:] As an alternative to the method presented, one could also train the abstracts to the independent claims. 
\end{itemize}

\section{Acknowledgment}
\noindent
We would like to thank Jochen Spuck, EconSight (Switzerland, Basel) and Carsten Guderian, PatentSight (Germany, Bonn), for their helpful and valuable discussions and support on patent data.

\newpage
\bibliography{FTO_Description_to_Claim}

\newpage

\appendix 
\section{Appendix}

\noindent\rule{\textwidth}{1pt}

\noindent
{\bf Reference patent 1: AU2012315252B2}

Abstract: An image browsing method, system and computer storage medium are provided. The image browsing method comprises the following steps: obtaining a user's operation request for an image (S110); comparing the operation request with a preset condition, and judging if it's required to buffer the image into a main buffer (S120); When it's required to buffer the image into the main buffer, further judging if rendering the image is first rendering (S130), a main thread buffers the image into the main buffer and performs the rendering by obtaining the image from the main buffer if it is (S140), and a main thread performs the rendering by obtaining an image adapted to the capacity of the main buffer from an image chain if it is not (S150); displaying the rendered image (S170). The image browsing method, system and computer storage medium avoid image browsing which is not smooth caused by a lot of time cost by generating the image required to be rendered, therefore the image can be browsed more smoothly.

\vspace{4mm}
\noindent
{\bf FTO patent 1.1: AU2012315252B2}

Claim 1: A method for image browsing, comprising: obtaining a user's operation request for an image; comparing the operation request with a preset condition, and determining whether it is required to cache the image into a main cache; when it is required to cache the image into the main cache, further determining whether it is the first time that the image is rendered; if yes, employing a main thread to cache the image into the main cache, to obtain the image from the main cache, and to render the image; otherwise, employing a main thread to obtain an image adapted to the capacity of the main cache from an image chain, and to render the image; and displaying the image rendered.

\vspace{4mm}
\noindent
{\bf FTO patent 1.2: US10186068B2}

Claim 1: A method of rendering an image, the method comprising: 
assigning a multi-threaded rendering unit to render a region of the image using a plurality of threads, the plurality of threads comprising at least a first thread and a second thread; 
for a first thread rendering a pixel on a scan line of the region, determining using an identifier of the first thread, an index of a start pixel on said scan line rendered by the second thread; 
rendering the pixel by the first thread using a position of the pixel determined based on the index of the start pixel and the identifier of the first thread; and 
rendering the image by combining rendered pixels of the region based on positions of the rendered pixels..

\vspace{4mm}
\noindent
{\bf FTO patent 1.3: US20200058273A1}

Claim 1: A method of rendering an object model, comprising: 
modifying, by a first thread executing on a computer system that is coupled to a display screen, a current object model to generate a first new object model, the first new object model being associated with an application running on the computing system, being a representation of one or more objects to display on the display screen, and being stored at a first memory address; 
storing, by the first thread, a copy of the first new object model at a second memory address; 
updating, by the first thread, a reference to identify the copy of the first new object model; and 
rendering, by a second thread executing on the computer system, a renderable object model identified by the reference into a buffer that is converted into pixels on the display screen, the first thread being independent of the second thread.

\vspace{4mm}
\noindent
{\bf FTO patent 1.4: US10089711B2}

Claim 1: A computer-implemented method comprising: 
receiving a request to load a digital image object for rendering to a display device; 
generating a first object from the digital image object; and 
managing memory operations of the first object in a cache memory using a management object generated separately when the first object is generated that references only to the first object, the management object added to the cache memory before the first object is added to the cache memory, and removed from the cache memory together with the first object.

\vspace{4mm}
\noindent
{\bf FTO patent 1.5: US9501415B1}

Claim 1: A method comprising: 
receiving, by a processing device, a request for a plurality of images from a rendering application providing a graphical user interface (GUI) for image scrolling, the rendering application executing in a high level language virtual machine (HLL VM) running on a mobile device; 
sequentially retrieving, by the processing device, the plurality of images from a disk of the mobile device using a fetching thread; 
decoding, by the processing device, the plurality of images using a plurality of decoding threads operating in parallel prior to the plurality of images being stored in a heap memory of the HLL VM, each of the plurality of decoding threads being executed by a distinct one of a plurality of processor cores of the mobile device; 
adding, by the processing device, decoded versions of the plurality of images to a first image cache in a native memory managed by an operating system of the mobile device; 
receiving, by the processing device, a request for a subset of the decoded versions of the plurality of images, the request being issued by the rendering application; and 
adding, by the processing device, the subset of the decoded versions of the plurality of images to a second image cache in the heap memory managed by the HLL VM, the subset of the decoded versions of the plurality of images to be presented by the rendering application in a user view of the GUI, the rendering application to request additional decoded versions of the plurality of images from the first image cache when a user scrolls through images presented in the user view.

\vspace{4mm}
\noindent
{\bf FTO patent 1.6: US10643746B2}

Claim 1: A system comprising: 
a network communication interface to receive compressed image pixel data; 
an image cache memory to cache the compressed DICOM image pixel data and rendered images; 
one or more processors coupled to the network connection interface and the memory and configured to implement an image rendering pipeline to perform image rendering in response to opening a healthcare study for review, wherein the image rendering includes determining whether compressed image pixel data associated with an image selected for display is in an image cache memory, and if so, then  
fetching the compressed image pixel data from an image cache, the image pixel data representing a pre-rendered version of an image from a series in the healthcare study, 
decompressing the compressed image pixel data to obtain the image pixel data, 
performing one or more image processing operations on the image pixel data, 
generating a displayable image from the image pixel data that has undergone image processing, and 
rendering the displayable image for display with a viewer on the display, and if not, then downloading from a remote location a version of the displayable image for display with the viewer; 
a display coupled to the one or more processors to display the displayable image with the viewer.

\vspace{4mm}
\noindent
{\bf FTO patent 1.7: US20170243322A1}

Claim 1: A system for multiple-frame buffering for graphically rendering a data set of images, comprising: 
a data store comprising a non-transitory computer readable medium storing a program of instructions for the implementation of the sorted linked list; 
a processor that executes the program of instructions, the instruction comprising the following steps: 
rendering a first set of image data every frame on a display; and 
rendering a second set of image data every other frame on the display; 
wherein the first set of image data is sourced from an external non-volatile memory, and temporarily buffered in an internal dynamic memory every frame, and 
the second set of image data is sourced from the external non-volatile memory, and temporarily buffered in the internal dynamic memory in a first frame, 
and moved to and temporarily buffered into a second frame in a second frame.

\vspace{4mm}
\noindent
{\bf FTO patent 1.8: US10452223B2}

Claim 1: A system for presenting a hierarchy of data objects in a three-dimensional browsing interface comprising: 
a user input device; 
a display; 
a processor; and 
a non-transitory computer-readable medium storing instructions that, when executed by the processor, cause the system to:  
arrange a hierarchy of data objects in a three-dimensional content browsing interface on the display, wherein a hierarchal relationship is represented by parent data objects having a higher position in the hierarchy being orbited, in three dimensions, by child data objects having a lower position in the hierarchy; 
select, based on a zoom depth relative to a displayed parent object of the parent data objects displayed in the hierarchal relationship, based on a quantity of a plurality of child data objects of the displayed parent object, and based on one or more content types of content included in the plurality of child data objects, a flocking layout automatically arranging the plurality of child data objects in an orbit around the displayed parent object; 
display the plurality of child data objects orbiting around the displayed parent object using the selected flocking layout; 
receive a selection command to select a data object of the hierarchy of data objects; 
navigate the three-dimensional content browsing interface to center the selected data object in the three-dimensional browsing interface; 
interpret a command received from the user input device, as an orbit command for orbiting around the selected data object; and 
constrain the orbit based on a content type for the selected data object.

\vspace{4mm}
\noindent
{\bf FTO patent 1.9: US8965933B2}

Claim 1: A method comprising: 
generating, by one or more computing devices, a rendering of a first section of data and storing the rendering of the first section of data in a first data buffer; 
determining, by the one or more computing devices, a data field that, when rendered, straddles the rendering of the first section of data and a rendering of a second section of data that will be stored in a second data buffer, the data field, when rendered, including a first portion and a second portion, wherein determining that the data field straddles the renderings includes determining that the first portion of the data field, when rendered, is stored in the first data buffer and that the second portion of the data field, when rendered, will be stored in the second data buffer; 
generating, by the one or more computing devices, a rendering of the data field separately from the rendering of the first section of data and the rendering of a second section of data and storing a rendered first portion of the data field and a rendered second portion of the data field in a cache; and 
generating, by the one or more computing devices, the rendering of the second section of data, including at least storing the rendered second portion of the data field, as stored in the cache, in the second data buffer.

\vspace{4mm}
\noindent
{\bf FTO patent 1.10: US9792881B2}

Claim 1: A display method, comprising: 
receiving a request to display content associated with a primary memory buffer and a secondary memory buffer, the content further associated with one or more rendering restrictions; 
identifying a content consumer based on one or more of the content, the primary memory buffer and the secondary memory buffer; 
determining whether the content consumer satisfies all of the one or more rendering restrictions; 
rendering content from the primary memory buffer when it has been determined that the content consumer satisfies all of the one or more rendering restrictions; and 
rendering content from the secondary memory buffer when it has been determined that the content consumer violates one or more of the one or more rendering restrictions.

\newpage
\noindent\rule{\textwidth}{1pt}

\noindent
{\bf Reference patent 2: US9659410B2}

Abstract: An augmented reality system is provided and a method for controlling an augmented reality system are provided. The augmented reality system, for example, may include, but is not limited to a display, a memory, and at least one processor communicatively coupled to the display and memory, the at least one processor configured to generate image data having a first resolution at a first rate, store the generated image data in the memory, and transfer a portion of the generated image data having a second resolution to the display from the memory at a second rate, wherein the second rate is faster than the first rate and the second resolution is smaller than the first resolution. This dual rate system then enables a head-tracked augmented reality system to be updated at the high rate, reducing latency based artifacts.

\vspace{4mm}
\noindent
{\bf FTO patent 2.1: US9659410B2}

Claim 1: An augmented reality system, comprising: 
a display; 
a memory; and 
a first processor communicatively coupled to the display and memory, the first processor configured to generate image data having a first field of view and having a first resolution at a first rate, wherein the generated image data includes control pixels, the control pixels including location data corresponding to the image data; a second processor communicatively coupled to the first processor, the display and the memory, the second processor configured to:  
store the generated image data in the memory; and 
transfer a portion of the generated image data having a second field of view smaller than the first field of view and having a second resolution to the display from the memory at a second rate, wherein the second rate is faster than the first rate and the second resolution is smaller than the first resolution, 
wherein the first processor and the second processor are communicatively connected via a twenty-four bit data channel, and the first processor is configured to generate the image data where each pixel of the image data includes 8-bits of green data, 6-bits of red data, 6-bits of blue data and 4-bits of weighting data, the weighting data defining a transparency of each pixel.

\vspace{4mm}

\noindent
{\bf FTO patent 2.2: US9666108B2}

Claim 1: A processing system for a display, the processing system comprising: 
a display memory; 
a compression module comprising compression circuitry and configured to receive a first frame of display update data from a host processor coupled to the processing system, compress the first frame of display update data according to a predefined compression algorithm, and store the compressed first frame of display update data in the display memory; and 
a display driver module comprising display circuitry and configured to:  
transmit, responsive to determining that an entirety of the first frame was compressed and stored in the display memory, a confirmation signal to the host processor; 
transmit, responsive to determining that a first portion of the first frame that is less than the entirety of the first frame was compressed and stored in the display memory, an indication signal identifying a second portion of the first frame that was not compressed and stored; and 
update the display using the compressed first portion of the first frame from the display memory, and the second portion of the first frame of display update data received from the host processor.

\vspace{4mm}

\noindent
{\bf FTO patent 2.3: US10762713B2}

Claim 1: A method for generating an augmented reality experience without a physical marker, comprising the steps of: 
a) collecting at least two frames from a video stream and designating one of the frames as a first frame; 
b) preparing said at least two collected frames for analysis by a graphical processor of a device 
c) selecting features from said at least two collected frames for comparison; 
d) isolating points on a same plane as a tracked point in the first frame by a central processor of the device; 
e) calculating a position of a virtual object in a second frame of the said at least two collected frames in 2D by the central processor; 
f) collecting a next frame from the video stream, designating the second frame as the first frame and designating the next frame as the second frame; 
g) rendering the virtual object on a display of the device to provide the augmented reality experiences on a web browser by the graphical processor; and 
h) repeating the steps b)-g) continuously until terminated by a user; and 
wherein the step of isolating points on the same plane as the track point in the first frame by the central processor comprises steps of removing outlier data using random sample consensus, removing outlier data again using random sample consensus, and calculating homographic transformation using at least four points.

\vspace{4mm}

\noindent
{\bf FTO patent 2.4: US10803826B2}

Claim 1: A special-purpose hardware device comprising: 
an image signal processor that:  
receives at least one image frame captured by a camera device; and 
forwards the image frame directly to a processing component within the special-purpose hardware device without temporarily buffering the image frame in memory; 
an input-formatting component that receives computer-generated imagery; 
a blending component communicatively coupled to the image signal processor and the input-formatting component, wherein the blending component generates at least one mixed-reality frame by combining the computer-generated imagery with the image frame received from the camera device; 
at least one hardware-accelerated image-correction component that performs at least one image-correction procedure on the mixed-reality frame before the mixed-reality frame is displayed; and 
wherein the image signal processor mitigates motion-to-photon latency of a head-mounted-display system that displays the mixed-reality frame by avoiding accessing memory in connection with the image frame.

\vspace{4mm}

\noindent
{\bf FTO patent 2.5: US9978114B2}

Claim 1: A system for optimizing processing and display of datasets, the system comprising: 
a memory device; 
a user interface comprising a display device; and 
a graphics processing and optimization (GPO) computing device coupled with said memory device and communicatively coupled to said user interface, wherein said GPO computing device is configured to:  
store a dataset comprising at least one data point in said memory device; 
select the at least one data point to display on said display device based on a first display request signal received from said user interface; 
accelerate graphical processing of the dataset using one or more optimization algorithms, wherein said GPO computing device is configured to assign a worker process to execute on the at least one data point; 
convert a display image having any geometric shape associated with the at least one data point into a renderable graphics component, including selecting a least-pixel representation of the at least one data point that best fits the display image associated with the at least one data point from one or more least-pixel representations, wherein the least-pixel representation is selected from transformation data stored in the dataset in said memory device; and 
display a graphical representation of a first subset of the dataset including the renderable graphics component at a first display resolution on said display device.

\vspace{4mm}

\noindent
{\bf FTO patent 2.6: IN271214B}

Claim 1: A device comprising: a storage medium having a plurality of shared M-dimensional (MD) registers; and a processing unit to implement a set of operations to pack in each shared MD register one or more shader variables whose sum of components equals M.  
1. A device for varying packing and linking in graphics systems, the device comprising: a storage medium having at least one shared M-dimensional (MD) register, wherein the at least one shared MD register has at least one row, and each row has at least M slots; and a processing unit coupled to the storage medium and configured to implement a set of operations to: consecutively pack in a first row of the at least one shared MD register, one or more shader variables from a vertex shader output file whose sum of components equals M; detect whether the M slots of the first row are full after loading each of the one or more shader variables in the first row; continue to consecutively pack the first row upon detection that the M slots of the first row are not full; and transfer the one or more shader variables packed in the first row of the at least one shared MD register to a respective MD cache register in a vertex cache, upon detection that the M slots of the first row are full.

\vspace{4mm}

\noindent
{\bf FTO patent 2.7: US10176739B2}

Claim 1: A method for performing partial frame updating, the method comprising: 
receiving instructions to generate a current frame of graphical data in a processor of a computing device; 
comparing the instructions to generate the current frame of graphical data to instructions to generate a previously generated frame of graphical data to determine an updated portion of graphical data; 
generating the updated portion of graphical data; 
sending the updated portion to a self-refreshing display panel communicatively coupled to the computing device; 
identifying a first plurality of pixels in the display panel corresponding to the updated portion; 
determining a second plurality of pixels in the display panel to be refreshed based on a threshold; and 
refreshing only the first and second pluralities of pixels in the self-refreshing display panel for a refresh period of the display panel.

\vspace{4mm}

\noindent
{\bf FTO patent 2.8: US10930084B2}

Claim 1: An electronic device comprising: 
a display configured to display a three-dimensional image; 
a camera configured to photograph a real image; and 
a controller configured to generate an image signal based on the real image and augmented reality (AR) image data and to provide the image signal to the display, the controller comprising:  
a multi-view image generator configured to convert the AR image data into multi-view AR images; 
a graphics processor configured to compose each of the multi-view AR images with the real image to generate multi-view composition images; and 
a processor configured to control a multi-view virtual camera and the graphics processor, to convert the multi-view composition images into the image signal, and to provide the image signal to the display, 
wherein the display comprises a light-field display, and the light-field display comprises:  
a display panel comprising a plurality of pixels; and 
a micro-lens array on the display panel and comprising a plurality of micro-lenses.

\vspace{4mm}

\noindent
{\bf FTO patent 2.9: US9626799B22}

Claim 1: A method for dynamically displaying multiple virtual and augmented reality scenes on a single display, the method comprising: 
determining a set of global transform parameters from a combination of user-defined inputs, user-measured inputs, and device orientation and position derived from sensor inputs of a device; 
calculating a first projection from a first configurable function of the set of global transform parameters, a first context provided by a user, and a first context specific to a first virtual and augmented reality scene; 
rendering the first virtual and augmented reality scene with the calculated first projection on a first subframe of a display; 
calculating a second projection from a second configurable function of the set of global transform parameters, a second context provided by a user, and a second context specific to a second virtual and augmented reality scene; 
rendering the second virtual and augmented reality scene with the calculated second projection on a second subframe of the display; 
displaying the rendered first and second virtual and augmented reality scenes simultaneously on the single display; and 
based on a change of the device orientation, simultaneously adjusting the display of the rendered first and second virtual and augmented reality scenes.

\vspace{4mm}

\noindent
{\bf FTO patent 2.10: IN201947028682A}

Claim 1: An improved method for caching image data intended for use with a head- mounted, augmented, mixed or virtual reality display system (HMD) having a graphics processing unit (GPU), a holographic processing unit (HPU) with an on-chip cache, system memory, and a display, the GPU generating images that can include text, holographic objects and other visual elements that can be displayed on the display, the system memory for storing GPU images, and the HPU configured to perform late stage adjustments to correct GPU images for movement of the HMD that may occur during the image rendering process, and the cache for constructing an output image based on the GPU generated image and the late stage adjustments, and the HPU generating a late stage adjustment matrix that maps each pixel of the late stage adjusted output image to the corresponding pixel of the GPU image, the method comprising acts for: 
 performing an initial pre-fetch of one or more rows of the GPU image from system memory, the one or more rows being selected in the order needed to construct the output image;  
 writing the initially pre-fetched one or more rows of the GPU image into the cache in the order needed to construct the output image based on the late stage adjustment matrix; and  
 walking the initially pre-fetched one or more rows of the GPU image in the cache in a raster scan order of the output image and according to the adjustment matrix and outputting the output image. 

\vspace{4mm}

\newpage
\noindent\rule{\textwidth}{1pt}

\noindent
{\bf Reference patent 3: US9911174B2}

Abstract: An image processing pipeline may process image data at multiple rates. A stream of raw pixel data collected from an image sensor for an image frame may be processed through one or more pipeline stages of an image signal processor. The stream of raw pixel data may then be converted into a full-color domain and scaled to a data size that is less than an initial data size for the image frame. The converted pixel data may be processed through one or more other pipelines stages and output for storage, further processing, or display. In some embodiments, a back-end interface may be implemented as part of the image signal processor via which image data collected from sources other than the image sensor may be received and processed through various pipeline stages at the image signal processor.

\vspace{4mm}

\noindent
{\bf FTO patent 3.1: US9911174B2}

Claim 1: An apparatus, comprising: 
an image signal processor that comprises:  
one or more front-end pipeline stages that processes pixels at an initial rate of pixels per clock cycle; 
a scaler; and 
one or more back-end pipeline stages that processes pixels at a different rate of pixels per clock cycle that is less than the initial rate of pixels per clock cycle, wherein the one or more back-end pipeline stages process the pixels subsequent to the one or more front-end pipeline stages; 
the image signal processor, configured to:  
receive a stream of pixel data collected at an image sensor according to an initial data size for an image frame; 
process the stream of pixel data through the one or more front-end pipeline stages of the image signal processor; 
scale, by the scaler, the stream of pixel data to a data size that is less than the initial data size for the stream of pixel data; 
process the stream of scaled pixel data through the one or more back-end pipeline stages; and 
provide the stream of scaled pixel data processed through the one or more back-end pipeline stages for display of the image frame.

\vspace{4mm}

\noindent
{\bf FTO patent 3.2: US9412147B2}

Claim 1: An apparatus, comprising: 
a plurality of processing pipelines, wherein each processing pipeline of the plurality of processing pipelines is configured to:  
receive pixel data; and 
process the pixel data to generate formatted data; 
a functional unit configured to:  
receive formatted data from each processing pipeline of the plurality of pipelines; and 
combine the formatted data to generate display data; 
a memory configured to store the display data; 
wherein a first processing pipeline of the plurality of processing pipelines includes one or more buffers, wherein a first buffer of the one or more buffers is configured to:  
receive and store the display data from the functional unit in response to a determination that the first processing pipeline is inactive; and 
send the stored display data to the memory; and 
circuitry configured to disable a first clock signal coupled to at least one of the plurality of processing pipelines in response to a determination that an amount of data stored in the first buffer is greater than or equal to a first buffer threshold value and that an amount of data stored in the memory is greater than or equal to a first memory threshold value.

\vspace{4mm}

\noindent
{\bf FTO patent 3.3: US9898799B2}

Claim 1: An electronic device comprising: 
a memory configured to store image data; 
a processor, functionally connected with the memory and including at least one image processing module configured to process an image, the processor configured to:  
obtain image data from the processing of the image, and 
store volatile information that is temporarily obtained from the image during the image processing; and 
an image pipeline formed of a plurality of image processing modules configured to perform the image processing in stages, 
wherein the processor is further configured to:  
obtain the image data, which is processed for the image in stages, from the image pipeline, 
store the volatile information, which is obtained from at least one of the plurality of image processing modules, in the memory, and 
store the volatile information in a header of a storage format of the image data based on a hierarchical structure grouping similar information in different scale ratios by layers.

\vspace{4mm}

\noindent
{\bf FTO patent 3.4: US10013744B2}

Claim 1: An image processing method, comprising: 
receiving from an imaging sensor input image data arranged in a plurality of pixel data rows corresponding to a grid of sensor pixels; 
using a digital image processor for sampling said input image data to provide sampled image data having less pixel data rows than said plurality of pixel data rows; 
using said digital image processor for correcting image distortion in at least a portion of said sampled data to provide processed image data, said correcting comprising collectively processing pixel data rows sampled from M consecutive pixel data rows of said input image data and stored in a row buffer having a storage capacity of no more than K pixel data rows, wherein M is between 2 i-1K and 2iK, and wherein i is a positive integer; and
transmitting said processed image data to a display and/or a computer readable medium.

\vspace{4mm}

\noindent
{\bf FTO patent 3.5: IN201934037994A}

Claim 1: An image signal processor for generating a converted image based on a raw image provided by an image sensor, the image signal processor comprising: processing circuitry configured to store data corresponding to a plurality of lines of a received image in a line buffer; perform an image processing operation by filtering the data stored in the line buffer based on at least one filter; and divide the raw image into a plurality of sub-images and request the plurality of sub-images from a first memory in which the raw image is stored, such that the plurality of sub-images are sequentially received by the line buffer, a width of each of the plurality of sub-images being less than a width of the line buffer, and the plurality of sub-images being parallel to each other.

\vspace{4mm}

\noindent
{\bf FTO patent 3.6: US20200098079A1}

Claim 1: An image signal processor for generating a converted image based on a raw image provided by an image sensor, the image signal processor comprising: 
processing circuitry configured to 
store data corresponding to a plurality of lines of a received image in a line buffer; 
perform an image processing operation by filtering the data stored in the line buffer based on at least one filter; and 
divide the raw image into a plurality of sub-images and request the plurality of sub-images from a first memory in which the raw image is stored, such that the plurality of sub-images are sequentially received by the line buffer, a width of each of the plurality of sub-images being less than a width of the line buffer, and the plurality of sub-images being parallel to each other.

\vspace{4mm}

\noindent
{\bf FTO patent 3.7: US10255655B1}

Claim 1: An apparatus, comprising: 
texture sample circuitry configured to retrieve texel data from a texture storage element for use in processing pixels in a frame of graphics data; 
a processing pipeline configured to:  
operate on a set of pixels, wherein information for the set of pixels is received by the processing pipeline in parallel, wherein the pipeline is configured to serially process the set of pixels such that data for different ones of the pixels in the set is processed in different pipeline stages of the pipeline during one or more clock cycles; and 
store texel information retrieved by the texture sampling circuitry from the texture storage element for a particular pixel at a stage of the processing pipeline, such that the stored texel information is available to process a subsequently processed pixel at the stage without accessing the texture storage element to retrieve the stored texel data for the subsequent pixel.

\vspace{4mm}

\noindent
{\bf FTO patent 3.8: US9996894B2}

Claim 1: An image processing device configured to process image raw data with an application processor and a video pipeline distinct from the application processor, comprising: 
the application processor for outputting at least a parameter and at least an instruction based on default or user setting to a video pipeline interface unit between the application processor and the video pipeline; 
the video pipeline interface unit including:  
a shared memory for storing the at least one parameter; and 
an inter-processor communication circuit for passing the at least one instruction to the video pipeline and reporting how the at least one instruction is treated by the video pipeline to the application processor; and 
the video pipeline for accessing the shared memory and processing the image raw data according to the at least one parameter, and for carrying out an operation indicated by the at least one instruction or refusing to execute the operation indicated by the at least one instruction.

\vspace{4mm}

\noindent
{\bf FTO patent 3.9: US20180365796A1}

Claim 1: An image processing device in which an image processing section for configuring a pipeline by connecting a plurality of processing modules for performing predetermined processing on input data in series and performing pipeline processing by each of the processing module sequentially performing the processing is connected to a data bus and performs image processing on data read from a data storage section connected to the data bus via the data bus, 
wherein the image processing section includes 
an input/output module incorporated into the pipeline as a processing module configured to perform processing different from the processing to be performed by each of the processing modules, and 
wherein the input/output module outputs processed data obtained by performing the processing of a first processing module which is the processing module located at a stage previous to a position where the input/output module is incorporated into the pipeline to an external processing section outside the image processing section, via an external interface section for inputting and outputting data to and from the external processing section without involving the data bus, and outputs externally processed data input by the external processing section performing external processing on the processed data to a second processing module which is the processing module located at a stage subsequent to the first processing module in the pipeline via the external interface section without involving the data bus, 
wherein the external interface section converts data to be transmitted in a format according to a specification of the image processing section when pixel data is received from the input/output module into a format of pixel data to be processed by the external processing section, and 
wherein the external interface section converts a format of externally processed pixel data output from the external processing section into a format in which the image processing section performs image processing when the externally processed pixel data is transmitted from the external processing section.

\vspace{4mm}

\noindent
{\bf FTO patent 3.10: GB2469526B}

Claim 1: A graphics processing apparatus for generating a frame of pixel data values to be stored in a frame buffer memory, said graphics processing apparatus comprising: a graphics processing pipeline including: (i) pixel value generating circuitry responsive to one or more first input parameters to generate pixel values at a first resolution for a region within said frame; (ii) a pipeline memory coupled to said processing pipeline and configured to store said pixel values at said first resolution; and (iii) write back circuitry coupled to said pipeline memory' and configured to wiite pixel values stored in said pipeline memory' to said frame buffer memory'; wherein said graphics processing pipeline further includes: (iv) programmable resolving circuitry responsive to one or more graphics program instructions and one or more second parameters to read said pixel values at said first resolution from said pipeline memory'· and perform a resolving operation specified by said graphics program instruction and said one or more second parameters to generate pixel values at a second resolution to be written by said write back circuitry to said frame buffer memory', said second resolution being different to said first resolution; wherein at said first resolution said region within said frame is represented by a first number of pixels and at said second resolution said region within said frame is represented by a second number of pixels.

\newpage
\noindent\rule{\textwidth}{1pt}

\noindent
{\bf Reference patent 4: US10375379B2}

Abstract: A 3D display device is disclosed, which comprises: a display panel and a modulating unit disposed on the display panel and comprising plural columnar elements. The display panel comprises: a substrate; plural horizontal electrode lines disposed on the substrate and substantially arranged in parallel; and plural vertical electrode lines disposed on the substrate and substantially arranged in parallel, wherein the vertical electrode lines are interlaced with the horizontal electrode lines to define plural pixel units. In addition, the columnar elements are slanted at a slant angle to an extension direction of the vertical electrode line of the display panel, and the slant angle is in a range from 60 degree to 85 degree.

\vspace{4mm}

\noindent
{\bf FTO patent 4.1: US10375379B2}

Claim 1: A display device, comprising: 
a display panel, comprising:  
a substrate; 
plural horizontal electrode lines disposed on the substrate and substantially arranged in parallel; and 
plural vertical electrode lines disposed on the substrate and substantially arranged in parallel, wherein the vertical electrode lines are interlaced with the horizontal electrode lines to define plural pixel units; and 
a modulating unit disposed on the display panel and comprising plural columnar elements, 
wherein the columnar elements are slanted at a slant angle to an extension direction of the vertical electrode lines, and the slant angle is in a range from 70 degree to 80 degree, 
wherein a width of each of the pixel units along extension directions of the horizontal electrode lines is larger than a width of each of the pixel units along extension directions of the vertical electrode lines.

\vspace{4mm}

\noindent
{\bf FTO patent 4.2: US20150168613A1}

Claim 1: A display device comprising: 
an array of display elements; and 
an optical diffuser disposed forward of the array of display elements, the optical diffuser having a bottom surface facing the array of display elements, the optical diffuser comprising:  
a layer of filler material having a first index of refraction; and 
a plurality of spaced-apart protrusions having heights substantially perpendicular to the bottom surface and extending into the layer of filler material, at least some of the plurality of protrusions being optically transmissive and having varying heights, each of the plurality of protrusions having an index of refraction different from the first index of refraction.

\vspace{4mm}

\noindent
{\bf FTO patent 4.3: US9047839B2}

Claim 1: A liquid crystal display device comprising: 
an array of pixels each having a memory function; 
a driving section that supplies a common voltage to a counter electrode of a liquid crystal capacitor, and supplies one of a first voltage and a second voltage to a pixel electrode of the liquid crystal capacitor, the first voltage being the same as the common voltage, the second voltage reversing polarity every predetermined period; and 
an adjusting section that adjusts an amplitude of at least the second voltage, 
wherein the adjusting section adjusts a voltage value on only a positive side of at least one of the first voltage and the second voltage to become higher than a voltage value on the positive side of the common voltage.

\vspace{4mm}

\noindent
{\bf FTO patent 4.4: US10204393B2}

Claim 1: A system comprising: 
a display to output visual content; and 
a semiconductor package apparatus coupled to the display, the semiconductor package apparatus including:  
a substrate; and 
logic coupled to the substrate, the logic to:  
determine a position associated with one or more polygons in unresolved surface data corresponding to the visual content, 
select an anti-aliasing sample rate based on a state of the one or more polygons with respect to the position, and 
uniformly resolve the unresolved surface data at the position in accordance with the selected anti-aliasing sample rate, wherein the selected anti-aliasing sample rate is to vary across a plurality of pixels.

\vspace{4mm}

\noindent
{\bf FTO patent 4.5: US9208714B2}

Claim 1: A display panel, comprising: 
an active matrix pixel array comprising:  
a plurality of gate lines; 
a plurality of source lines; and 
a plurality of pixel elements arranged in a matrix, each pixel element being coupled to the corresponding gate line and the corresponding source line, each pixel element comprising:  
an image data storage capacitor for storing an image data; 
a sample unit having a control terminal for receiving a sample control signal; 
a capacitive element having a first terminal coupled to a pixel electrode of the image data storage capacitor via the sample unit; 
a first refresh unit having a control terminal coupled to the first terminal of the capacitive element; 
a second refresh unit having a control terminal for receiving a refresh control signal, the first and second refresh units being serially coupled with each other and between the corresponding source line and the image data storage capacitor for receiving a data signal; and 
a shunt unit having a control terminal coupled to the pixel electrode of the image data storage capacitor, a data terminal coupled to the first terminal of the capacitive element, and another data terminal for receiving a shunt control signal, wherein each of the shunt control signal and the data signal sequentially has a plurality of voltages during a plurality of period, and the voltages are in a monotonic order, wherein each of the data signal and the shunt control signal sequentially has a first voltage during a first period, a second voltage during a second period, a third voltage during a third period, and a fourth voltage during a fourth period, these units include N-type transistors, the first voltage is higher than the second voltage, and wherein the transition from the first voltage to the second voltage in the shunt control signal is in advance of the transition from the first voltage to the second voltage in the data signal; 
a source driver for driving the source lines; and 
a gate driver for driving the gate lines.

\vspace{4mm}

\noindent
{\bf FTO patent 4.6: US20150287393A1}

Claim 1: A method for driving an organic light emitting display device, comprising: 
storing, during a period of at least one normal frame, a voltage corresponding to a data signal in a pixel; and 
driving, during a period of k (k is a natural number of one or more) continuous frames arranged continuously after the period of the normal frame, the pixel in accordance with the data signal stored during the period of the at least one normal frame.

\vspace{4mm}

\noindent
{\bf FTO patent 4.7: IN201647038229A}

Claim 1: A method for coarse pixel shading (CPS) comprising: 
 pre-processing a graphics mesh by creating a tangent-plane parameterization of desired vertex attributes for each vertex of the mesh; and  
 performing rasterization of the mesh in a rasterization stage of a graphics pipeline using the tangent-plane parameterization.  
1. A method for coarse pixel shading (CPS) comprising: pre-processing a graphics mesh by creating a tangent-plane parameterization of desired vertex attributes for each vertex of the mesh; and performing rasterization of the mesh in a rasterization stage of a graphics pipeline using the tangent-plane parameterization.

\vspace{4mm}

\noindent
{\bf FTO patent 4.8: US10522113B2}

Claim 1: An apparatus comprising: 
one or more substrates; and 
logic coupled to the one or more substrates, wherein the logic is implemented in one or more of configurable logic or fixed-functionality logic hardware, the logic coupled to the one or more substrates to: 
identify a pixel location with respect to a plurality of display planes, 
store image data associated with the pixel location and the plurality of display planes to adjacent memory locations, 
simultaneously render the image data from the adjacent memory locations across the plurality of display planes, 
render a source view associated with a first display plane in the plurality of display planes, 
re-project the rendered source view to a second display plane in the plurality of display planes to obtain a re-projected view, and 
fill one or more holes in the re-projected view based on one or more of extended field of view data corresponding to the source view, rasterization data corresponding to the source view or rasterization data corresponding to the re-projected view, 
wherein when the one or more holes are filled based on the rasterization data corresponding to the source view, and wherein the logic coupled to the one or more substrates is to: 
disable a first depth test during rendering of the source view, and 
conduct the first depth test during filling of the one or more holes in the re-projected view.

\vspace{4mm}

\noindent
{\bf FTO patent 4.9: IN201403134I4}

Claim 1: A method comprising: presenting an object via a display operatively coupled with an electronic device; presenting a first visual indicator that relates time information associated with a motion of the object with a motion path of the object; and presenting a time line associated with the time information.

\vspace{4mm}

\noindent
{\bf FTO patent 4.10: US20170076417A1}

Claim 1: An integrated circuit, comprising: 
a memory; and 
display pipeline circuitry coupled to the memory and configured to:  
produce a sequence of frames for a display device, wherein the sequence includes a first frame and a second, subsequent frame; 
identify pixels of the second frame that differ from pixels of the first frame; and 
transmit, to the display device, content of the identified pixels and a bitmap indicating the locations of the identified pixels within the second frame.

\newpage
\noindent\rule{\textwidth}{1pt}

\noindent
{\bf Reference patent 5: US20150302541A1}

Abstract: [Problem] To provide an image data distribution server that is capable of distributing digitally watermarked image data that correspond to the ranks of user access rights to client terminals. [Solution] An image data distribution server (11) has: an image data generation means that generates multiple sets of image data of various kinds on individual images of multiple pieces of electronic information of various kinds that are stored in the server; a rank-based classification means that, upon receiving requests from users for distribution of the image data, classifies the access rights of the respective users to the image data in terms of ranks on the basis of authentication results for the users; a display form changing means that adds digital watermarks that correspond to the respective access right ranks classified by the rank-based classification means to the image data in an unremovable manner; and an image data distribution means that distributes the multiple sets of image data of various kinds, including the image data to which the digital watermarks are added, to client terminals (12A-12C).

\vspace{4mm}

\noindent
{\bf FTO patent 5.1: US20150302541A1}

Claim 1: An image data distribution server distributing various types of a plurality of pieces of image data to clients via the Internet in response to requests from a plurality of users, the image data distribution server comprising: 
image data creating means for creating various types of a plurality of pieces of image data which are individual images formed from various types of a plurality of pieces of electronic information stored in the image data distribution server; 
rank-based classifying means for, when the users request distribution of the image data via the clients, classifying access authority of the users for the image data into ranks based on authentication results of the users; 
display form changing means for changing display forms of the image data according to ranks of the access authority classified by the rank-based classifying means; and 
image data distributing means for distributing the various types of a plurality of image data including the image data for which the display forms are changed by the display form changing means to the clients.

\vspace{4mm}

\noindent
{\bf FTO patent 5.2: US9734549B2}

Claim 1: A method, carried out by a memory controller, the method comprising: arbitrating access to at least a portion of unified memory among a plurality of client data access requests from a plurality of clients to a plurality of memory channels, wherein the plurality of memory channels are accessible bypassing arbitration of the plurality of client data access requests to said at least a portion of the unified memory in response to a CPU data access request to said at least same portion of the unified memory; and controlling the plurality of memory channels to access client data address space and CPU data address space of the same unified memory simultaneously.

\vspace{4mm}

\noindent
{\bf FTO patent 5.3: US9530177B2}

Claim 1: A data access method, for a data access device to access data from N layers to display an image, N being a positive integer, each of the N layers comprising a horizontal start point, a horizontal end point, a vertical start point and a vertical end point, the method comprising: 
dividing the image into a plurality of regions according to the horizontal start points, the horizontal end points, the vertical start points and the vertical end points, wherein the regions respectively correspond to the layers; and 
accessing data from the respective layers corresponding to the regions when displaying the image, 
wherein the steps of dividing the image into the plurality of regions and accessing data from the respective layers corresponding to the regions comprise: 
defining a plurality of vertical sections according to the N vertical start points and the N vertical end points, and performing a sorting procedure according to a layer display priority sequence for one or a plurality of layers included in a range of each of the vertical sections to obtain a vertical sequence result for each of the vertical sections, where N is a positive integer; and 
defining a plurality of horizontal sections according to the N horizontal start points and the N horizontal end points, and performing a sorting procedure according to a layer display priority sequence and one or a plurality of layers included in a range of each of the horizontal sections to obtain a horizontal sequence result for each of the horizontal sections; 
wherein, the vertical sections are for the data access device to determine whether a vertical position of a horizontal scan line is located in one of the vertical sections; when the vertical position of the scan line is located in one of the vertical sections, the vertical sequence result corresponding to the vertical section and the N horizontal sequence results are for the data access device to determine to which of the N layers the horizontal section in the vertical section corresponds, and are for the data access device to accordingly access data from one of the N layers.

\vspace{4mm}

\noindent
{\bf FTO patent 5.4: US9451041B2}

Claim 1: A computer-implemented method for processing geographic content in client computing devices, the method comprising: 
receiving, at a client computing device from a server, geographic data and a caching policy for the geographic data, the geographic data related to a first geographic location and the caching policy defining one or more conditions to cache or discard the geographic data at the client computing device including a distance threshold; 
displaying the geographic data within a viewport of the client computing device when the viewport depicts a digital map corresponding to the first geographic location; 
moving the viewport to an updated position corresponding to a second geographic location; 
determining a physical distance between the first geographic location and the second geographic location; 
comparing the determined physical distance to the distance threshold; and 
caching or discarding the received geographic data based at least in part on the comparison, including caching the received geographic data when the physical distance is below the distance threshold and discarding the received geographic data when the physical distance is above the distance threshold.

\vspace{4mm}

\noindent
{\bf FTO patent 5.5: US20170083450A1}

Claim 1: A method comprising: 
using a page table to convert an address in order to access data stored in memory; and 
converting said data according to information stored in the page table.

\vspace{4mm}

\noindent
{\bf FTO patent 5.6: US10719433B2}

Claim 1: A method of storing and accessing data comprising the steps of: 
generating in a computer readable storage medium a first array comprising elements representing a variable sized rectangle of image pixels and containing information about transform settings and coefficients representing those pixels; 
generating in the computer readable storage medium a second array comprising elements representing a variable sized rectangle of image pixels and containing information about prediction and coding modes of those pixels; and 
generating in the computer readable storage medium a third array comprising elements representing a small fixed size square of image pixels and containing two indices, a first index specifying an element of the first array that covers the corresponding pixels and a second index specifying an element of the second array that is associated with the corresponding pixels.

\vspace{4mm}

\noindent
{\bf FTO patent 5.7: US9667531B2}

Claim 1: An image processing apparatus comprising: 
a storage configured to store image data; and 
a processor configured to,  
determine a transfer path via which the image data is to be transferred from among multiple transfer paths connected to the storage, 
select the transfer path via which the image data is to be transferred based on the determined transfer path, 
determine whether the determined transfer path is changed from a transfer path used in most recent transfer of the image data, 
if the processor determines that the determined transfer path is changed from the transfer path used in the most recent transfer of the image data, wait for responses to all of already-issued access requests to the storage and issue a next access request to the storage after waiting for the responses to the all of already-issued access requests to the storage, and 
if the processor determines that the determined transfer path is not changed from the transfer path used in the most recent transfer of the image data, issue a next access request so as to control a difference between a number of already-issued access requests and a number of already-received responses.

\vspace{4mm}

\noindent
{\bf FTO patent 5.8: CA2813877C}

Claim 1: A computer implemented method for processing a digital image by an electronic device comprising a display and an input device, said method comprising: maintaining ordered cached digital images arranged in a predetermined order, said ordered cached digital images comprising a base digital image and a subsequent plurality of modified cached digital images, each modified cached digital image associated with a different image modification category, wherein each subsequent modified cached digital image is generated by performing image modification operations relating to its respective image modification category on the immediately preceding cached image in the list; and receiving a user request to add a category of image modification and adding a cached image corresponding to the added image modification category to the maintained cached images in accordance with the predetermined order.

\vspace{4mm}

\noindent
{\bf FTO patent 5.9: US20170140497A1}

Claim 1: A display operation system in which a server apparatus is connected to a client apparatus and an operation is performed with a monitor screen of the client apparatus, wherein 
data for use in individual and independent control in the server apparatus is stored and loaded in the server apparatus, 
data for use in individual and independent control in the client apparatus is stored and loaded in the client apparatus, and 
data to be shared and for use in consolidated control over the entire display operation system is shared by the server apparatus and the client apparatus.

\vspace{4mm}

\noindent
{\bf FTO patent 5.10: US9153211B1}

Claim 1: A computer-implemented method for tracking accesses to virtual addresses, the method comprising: 
Initializing a virtual access bit buffer that includes a plurality of virtual access bits by clearing each virtual access bit in the plurality of virtual access bits, wherein the virtual access bit buffer includes a different virtual access bit for each virtual page residing within a virtual memory space allocated for a graphics context; 
receiving a virtual address from a client conveying a memory access request associated with the graphics context; 
matching a virtual page associated with the virtual address with a page table entry; 
deriving a physical address from a physical page contained in the page table entry; 
determining that virtual access tracking is enabled; and 
setting a virtual access bit located in the virtual access bit buffer, wherein the virtual access bit corresponds to the virtual page.

\end{document}